\documentclass[wcp]{jmlr}

\usepackage{longtable}

\usepackage{booktabs}
\usepackage{hyperref}
\usepackage{url}
\usepackage{xcolor}

\usepackage{graphicx}
\usepackage{booktabs}

\usepackage{amsmath}
\usepackage{newtxmath}
\usepackage{multirow}
\usepackage{float}
\usepackage{soul}
\usepackage{comment}
\usepackage{ragged2e}

\usepackage{lineno}

\makeatletter
\let\Ginclude@graphics\@org@Ginclude@graphics 
\makeatother

\jmlrvolume{222}
\jmlryear{2023}
\jmlrworkshop{ACML 2023 Preprint}

\title[MLET]{Enhancing Cross-Category Learning in Recommendation Systems with Multi-Layer Embedding Training}

 \author{\Name{Zihao Deng} \Email{zihaodeng@utexas.edu}\\
  \Name{Benjamin Ghaemmaghami} \Email{ben.ghaem@utexas.edu}\\
  \addr The University of Texas at Austin
  \AND
  \Name{Ashish Kumar Singh} \Email{ashish101@gmail.com}\\
  \addr E2open
  \AND
  \Name{Benjamin Cho} \Email{bjcho@utexas.edu}\\
  \Name{Leo Orshansky} \Email{orshaleo@utexas.edu}\\
  \Name{Mattan Erez} \Email{mattan.erez@utexas.edu}\\
  \Name{Michael Orshansky} \Email{orshansky@utexas.edu}\\
  \addr The University of Texas at Austin}

\editors{Berrin Yan{\i}ko\u{g}lu and Wray Buntine}

\begin{document}

\maketitle

\begin{abstract}
Modern DNN-based recommendation systems rely on training-derived embeddings of sparse features. Input sparsity makes obtaining high-quality embeddings for rarely-occurring categories harder as their representations are updated infrequently. 
We demonstrate a training-time
technique to produce superior embeddings via effective cross-category learning and theoretically explain its surprising effectiveness. 
The scheme, termed the multi-layer embeddings training (MLET), trains embeddings using factorization of the embedding layer, with an inner dimension higher than the target embedding dimension. 
For inference efficiency, MLET converts the trained two-layer embedding into a single-layer one thus keeping inference-time model size unchanged.

Empirical superiority of MLET is puzzling as its search space is not larger than that of the single-layer embedding. The strong dependence of MLET on the inner dimension is even more surprising. 
We develop a theory that explains both of these behaviors by showing that MLET creates an adaptive update mechanism modulated by the singular vectors of embeddings.
When tested on multiple state-of-the-art recommendation models for click-through rate (CTR) prediction tasks, MLET consistently produces better models, especially for rare items. 
At constant model quality, MLET allows embedding dimension, and model size, reduction by up to 16x, and 5.8x on average, across the models. 

\end{abstract}

\section{Introduction}
Recommendation models (RMs) underlie a large number of applications and improving their performance is increasingly important. The click-through rate (CTR) prediction task is a special case of general recommendation that seeks to predict the probability of a user clicking on a specific category. User reactions to earlier-encountered instances are used in training a CTR model and are described by multiple features that capture user information (e.g., age and gender) and category information (e.g., movie title, cost)~\cite{Ouyang2019ClickthroughRP}. Features are either numerical or categorical variables. 
A fundamental aspect of modern recommendation models is their reliance on embeddings which map categorical variables into dense representations in an abstract real-valued space.
State-of-the-art RMs increasingly use deep neural networks. Most high-performing models use a combination of multi-layer perceptrons (MLPs) to process dense features, linear layers to generate embeddings of categorical features, and either dot products or sub-networks that generate higher-order interactions. The outputs of the interaction sub-networks and MLPs are used as inputs into a linear (logistic) model to produce the CTR prediction. 
Broadly, the above describes modern deep recommender systems, including: DLRM~\cite{Naumov2019DeepLR}, Wide and Deep (WDL)~\cite{WideDeep}, Deep and Cross (DCN)~\cite{Wang2017DeepC}, DeepFM~\cite{Guo2017DeepFMAF}, 
Neural Factorization Machine (NFM)~\cite{he2017NFM}, AutoInt~\cite{Song2019AutoIntAF}, and xDeepFM~\cite{Lian2018xDeepFMCE}. 
\vspace{0cm}

\begin{figure}[htbp]
	\subfigure{
		\includegraphics[width=7cm,height=7cm]{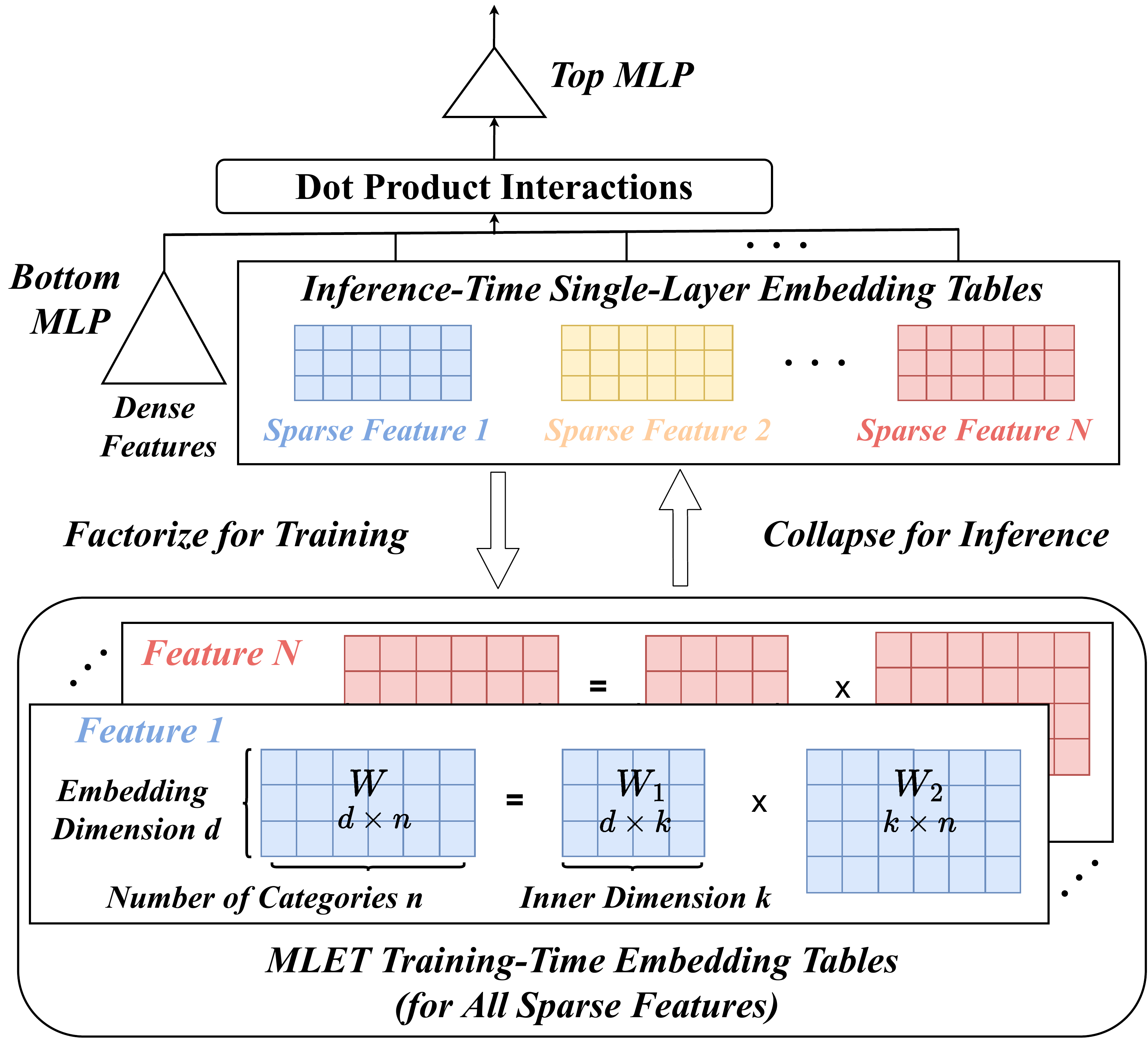}
		\label{fig:mlet_overview_diagram}
        }
        \subfigure{
		\includegraphics[width=7cm,height=6cm]{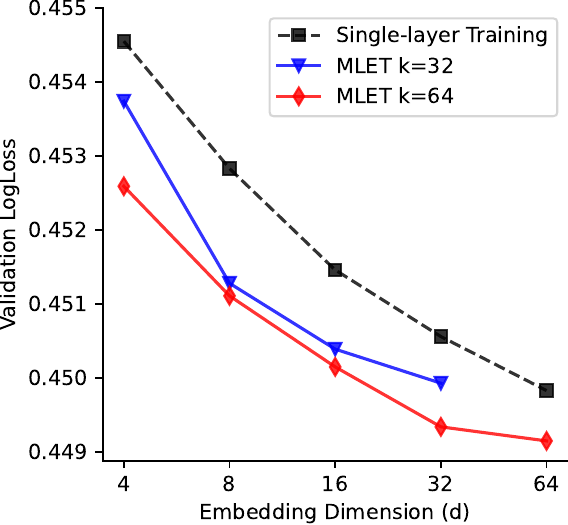}
		\label{fig:tradeoff}
        }
    \caption{(a) MLET trains a two-layer model but collapses it to a single layer for inference. (b) MLET results in superior performance compared to single-layer embedding training. Importantly, improvement grows with MLET's inner dimension $k$.}
    \vspace*{-0.5cm}
	\label{fig:1}
\end{figure}
\label{sec:MLET_better}
\vspace{0cm}
\emph{We propose and study an embarrassingly simple overparameterization technique for enhancing embeddings training by enabling effective cross-category learning, the MLET.}
\label{ssec:MLET_DEF}
Figure~\ref{fig:1}(a) illustrates the technique and the transformations involved.
Let the full embedding table be $W \in \mathbb{R}^{d \times n}$, where $n$ is the number of elements in the table and $d$ is the embedding dimension. Each column of the table represents the embedding of a category. 
The conventional way of training $W$ is to represent it by a single linear layer and train it jointly with the rest of recommendation models.
MLET uses a two-layer architecture that factorizes the embedding table $W$ in terms of  $W_1$ and $W_2$: $W = W_1  W_2\,\,\text{ with }W_1 \in \mathbb{R}^{d \times k}\,,W_2 \in \mathbb{R}^{k \times n}$.
$k$ is a hyperparameter representing the inner dimension of embedding factorization.
Vector $q\in \mathbb{Z}^{n}$ denotes a one-hot encoding of a query to $n$ categories. 
The embedding lookup is represented by a matrix-vector product: $r = W_1W_2q$.
$r\in \mathbb{R}^{d}$ denotes the embedding of the queried category. 
In MLET, $W_1$ and $W_2$ are only used during training: after training, only their product $W = W_1W_2$ is retained. This reduces a two-layer embedding into a single-layer one for inference.

{\color{black}



The contributions of this paper are two-fold.
First, we empirically show MLET's effectiveness in improving performance and reducing model size. Tested on seven state-of-the-art recommendation models with two public CTR datasets, MLET allows a reduction of 16x (5.8x less on average) in inference-time embedding parameters compared to single-layer embedding training at constant performance.  
Figure ~\ref{fig:1}(b) compares, using DLRM model on the Criteo-Kaggle dataset~\cite{CriteoKaggle2014}, the quality-size trade-offs obtained by a conventional single-layer embedding training and MLET, showing MLET's sizeable benefits.

More importanlty, we present a theory to explain the puzzling effectiveness of MLET. 
Because MLET does not increase the search space of the single-layer embedding, nor does increasing the inner dimension enlarges the search space when the inner dimension is bigger than the embedding dimension, two aspects of MLET seem surprising. 
The first is why it is superior to single-layer embedding training. The second is why its quality continues to improve with a larger inner dimension, which is already bigger than the embedding dimension.
{\color{black}To answer the first question,} 
we point out that in each iteration of conventional single-layer embedding training, only the embeddings corresponding to the queried categories get updated. Due to sparsity of queiries, only a small fraction of embeddings are updated. Importantly, the rarely-occurring categories are updated less frequently, compared to the more frequent ones. \emph{In contrast, MLET leads to embeddings of all categories being updated on each training iteration.} Effectively, knowledge from the queried categories is used to also update embeddings of non-queried categories. We call this behavior 
\emph{cross-category learning}. As we observe empirically, cross-category learning 
leads to much more effective learning, especially, for rarely-occurring categories. 
{\color{black}
To answer the second question, we present a theory that identifies the source of cross-category learning as due to a reweighting mechanism created by MLET.
In every training iteration, the reweighting factor uses the singular values of MLET’s embedding layers to measure the agreement between the update (that is equal to the update that would take place in the single-layer model) and the already-learned embeddings. It then boosts/attenuates updates in directions that agree/disagree with the learned embeddings.
The number of non-zero reweighting factors explains why MLET's performance shows a clear dependence on its inner dimension.
}  
}

\vspace{0cm}
\section{Related Work}
There are three relevant threads of related work: (1) experimental investigations of overparameterization techniques, (2) theoretical aspects of overparameterization, and (3) table-compression and table-decomposition approaches.


Multiple authors conducted experimental work proposing overparameterization techniques to enhance training performance.
E.g., \cite{Yang2022ExpansionSqueezeBL,guo2020expandnets} show that overparameterization leads to enhancement of performance and generalization in the context of CNNs.
On the theory side, \cite{Arora2019ImplicitRI} developed a theory of overparameterization in deep linear neural networks, with the primary mechanism being a tendency towards lower rank that improves generalization. However, the theory does not appear to be helpful in understanding the behavior of embedding layers in the context of complete RMs. 
The predicted tendency towards low rank is not observed empirically, nor would it explain the observed faster training loss convergence.
Moreover, these prior theoretical frameworks \cite{Arora2018OnTO,Arora2019ImplicitRI} do not explain why the superiority of overparameterization is related to the amount of overprameterization, which is experimentally observed both by \cite{Yang2022ExpansionSqueezeBL,guo2020expandnets}  and by our MLET method.

{
The benefit of MLET is in producing embedding tables with superior performance for fixed table size. An orthogonal set of approaches for achieving this goal includes {\color{black}compression via post-training} pruning and quantization~\cite{ling-etal-2016-word,nearlossless_tissier,sun2016sparse}; {\color{black}training-aware} pruning and quantization~\cite{alvarez2017compression,naumov2018periodic}; and hashing tricks that share embeddings within or between tables~\cite{attenberg2009collaborative,shi2019compositional}. Techniques that utilize statistical knowledge of embedding usage (access frequency) have also been developed to adapt the embedding dimension or precision to usage, with more-compact representations of less-accessed embeddings~\cite{ginart2019mixed,yang2020mixedprecision}. 
In addition to being an orthogonal approach to those above, MLET produces high-quality embeddings without assuming any prior knowledge of access frequency and without reducing parameter precision. 
Instead, under the same inference-time embedding size, {\color{black} it achieves better performance by promoting more frequent and more informative updates of embeddings than those in single-layer training.}
}

{
{\color{black}We also highlight MLET's critical differences to some decomposition techniques.} For example, trained embedding tables can be compressed via a low-rank SVD approximation~\cite{bhavana2019block} or using a tensor-train decomposition~\cite{khrulkov2019tensorized}. 
TT-Rec~\cite{Yin2021TTRecTT} uses tensor-train decomposition to represent embeddings and is similar to MLET in that multiple tensors instead of one are used in learning each embedding table.
{\color{black}However, TT-Rec and low-rank SVD are orthogonal to MLET, and completely differ in their working mechanism from it.
Both TT-Rec and low-rank SVD utilize \emph{underparameterization} to maintain training performance.
MLET, in contrast, employs \emph{overparameterization} to improve training performance.}
With the empirical benefits demonstrated by MLET, we believe many opportunities are now open for exploring the combinations of MLET with the above techniques.

We conclude this survey by pointing out that no other work has shown how to enhance cross-category training or theoretically analyzed its mechanism.
{\color{black}\emph{Our research is pioneering in that it formally explores the advantages of overparameterization within the realm of recommendation models. Our novel theoretical framework stems from a rigorous analysis of gradient flow and its impact on the evolution of embeddings. Significantly, this newly developed theory elucidates not only the empirical benefits associated with overparameterization but also expounds the correlation between the degree of overparameterization and the consequent enhancement of training performance.}
}
}

\section{{\color{black}Breaking The Sparsity of Embedding Updates}}

\subsection{{\color{black}Cross-Category Learning in MLET}}
\label{sec:321}

Consider MLET's embedding $W=W_1W_2$.
{
The factorization explicitly formulates the embeddings as linear combinations of the embedding basis formed by the columns in $W_1$. 
}
Let the loss function be $L$ and loss gradient $G=\frac{\partial L}{\partial w}$. 
Given a learning rate $\eta$, for training of the single-layer model, embedding updates are
\begin{equation}
    W = W - \eta G
    \label{eq:update1}
\end{equation}
$G$ is sparse, with only one column being non-zero.
To see this, let $g$ be the gradient of loss w.r.t. $r$, i.e., $g=\partial L/\partial r$. 
Notice that $q$ is a one-hot encoded vector representing the queried category, whose embedding is $r\in \mathbb{R}^{d}$. We use $C$ to denote the index of the queried category.
By chain rule, one can show that $G=gq^T$. Therefore, only the $C^\mathit{th}$ column of G is non-zero and is equal to $g$.
With batch size $b>1$, the conclusion extends: no more than $b$ columns are non-zero.
Because $b<<n$ in embedding tables, G is still highly column-sparse.

\begin{flushleft}
In contrast, with the same learning rate, embedding updates of MLET are:
\end{flushleft}
\begin{equation}
    W = W - \eta W_1W_1^TG - \eta GW_2^TW_2
    \label{eq:update2}
\end{equation}
The derivation is as follows. First, let $W(t)$ be the embedding at $t^\mathit{th}$ iteration and
$G$ be the gradient of embedding: $G=\partial L /\partial W(t)$.
The equivalent single-layer embedding updates of MLET are:
\begin{equation}
\begin{split}
    W(t+1)&= W_1(t+1)W_2(t+1) 
    = \left(W_1(t) - \eta \frac{\partial L}{\partial W_1(t)}\right)\left(W_2(t) - \eta \frac{\partial L}{\partial W_2(t)}\right)
\end{split}
\label{apdx:1}
\end{equation}
Applying similar analysis as the derivation of $G$, we have:
\begin{equation}
    \frac{\partial L}{\partial W_1} = gq^TW_2^T = GW_2^T,\,\,\frac{\partial L}{\partial W_2} = W_1^Tgq^T = W_1^TG
    \label{apdx:W1}
\end{equation}
Bringing Eq.\ref{apdx:W1} into Eq.\ref{apdx:1}, replacing $W_1(t)W_2(t)$ with $W(t)$, and ignoring the $O(\eta^2)$ term (following the convention of gradient flow analysis) lead to Eq.\ref{eq:update2}.
\vspace{0cm}
\begin{figure}[H]
	\centering
		\includegraphics[width=14cm]{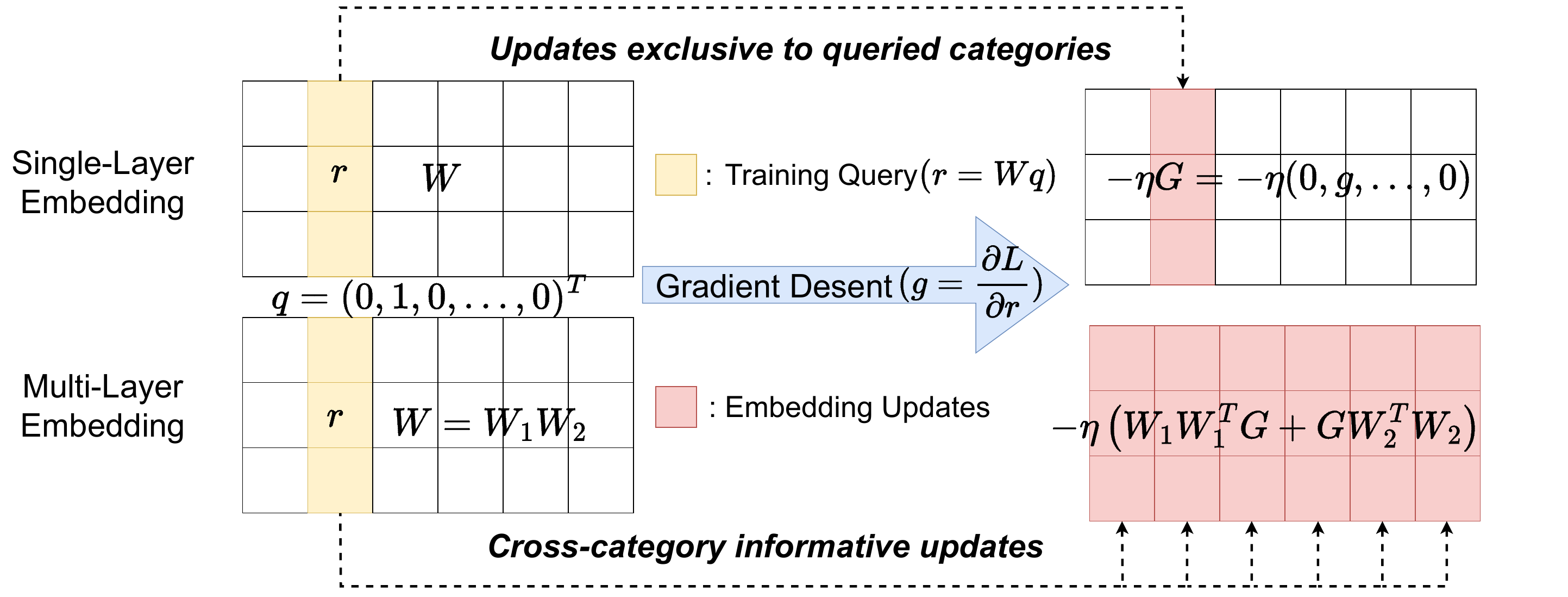}
    \vspace*{-0.2cm}
    \caption{MLET: cross-category informative updates. Knowledge of queried items is used to update all embeddings.} 
    \vspace*{-0.7cm}
	\label{fig:sparse2dense}
\end{figure}
\vspace{0cm}
{\color{black}
Comparing Eq.\ref{eq:update1} and Eq.\ref{eq:update2}, we observe that (Fig.~\ref{fig:sparse2dense}) in each step of single-layer embedding training, only one column of $W$ is updated, however, the whole embedding table $W$ is updated in MLET, because right-multiplying $G$ by a dense matrix $W_2^TW_2$ breaks its sparsity.
\emph{The dense updates mean that the information of queried categories in each training sample is also used to update other non-queried categories.
This property does not present in conventional single-layer embedding training and we refer to it as cross-category informative updates.}
}

\vspace{0cm}
\subsection{\color{black}{Reweighting of Embedding Updates}} 
\label{subsec:whyHelp}

\emph{Why does breaking the sparsity in this way help embedding training and what is the underlying working mechanism of MLET's cross-category learning?}
We present a theory that reformulates the embedding updates of two methods and pins down the difference to a term that reweights different embedding directions.


We introduce the following notation: vec($X$) represents the vectorization of the matrix X, formed by stacking the columns of X into a single column vector. 
$\otimes$ represents the Kronecker product operator.
The SVDs of $W_1$ and $W_2$ are denoted by $W_1 = U\Sigma_1X^T$ and $W_2 = Y\Sigma_2V^T$, and
$u_i$ and $v_j$ represent the $i^\mathit{th}$ column of $U$ and $j^\mathit{th}$ column of $V$, respectively.
{\color{black}Note that $i\in\{1,2,..d\}$ and $j\in\{1,2,..n\}$.}
We use $\sigma_1(i),\sigma_2(j)$ to denote the $i^\mathit{th}$ singular value in $\Sigma_1$ and $j^\mathit{th}$ singular value in $\Sigma_2$.
{\color{black}
For $i,j>k$, $\sigma_1(i),\sigma_2(j)$ are zeros.
}
We make the following claims.
\vspace{-0.1cm}
\begin{flushleft}\textbf{Claim 1.}$S=\{v_j\otimes u_i,i\in\{1,..d\},j\in\{1,..n\}\}$ is an orthornormal basis in $\mathbb{R}^{nd}$.\end{flushleft}  

\vspace{-0.5cm}
\begin{flushleft}\textbf{Claim 2.} There exists a set of $g_{ij}$ with $i\in\{1,..d\}$ and $j\in\{1,..n\}$ such that\end{flushleft}  
\begin{equation}
\mathrm{vec}(G)=\sum\limits_{i,j}g_{ij}v_j\otimes u_i
\label{eq:update4}
\end{equation}
To derive \textbf{Claim 1}, one can use $(v_j\otimes u_i)^T \cdot (v_q\otimes u_p) =  (v_j^Tv_q)\otimes(u_i^Tu_p)$ to prove that product of any vector in $S$ to itself is 1 and product of any two different vectors in $S$ is 0.
\textbf{Claim 2} follows directly from \textbf{Claim 1} and the fact that vec($G$) is in $\mathbb{R}^{nd}$.
Based on the above claims, we introduce our main theorem.
\vspace{-0.2cm}
\begin{flushleft}\textbf{Theorem 1. (Main Theorem)}
The embedding updates of the conventional single-layer training and those of MLET can be represented in basis $S$:\end{flushleft} 
\vspace{0cm}
\begin{equation}
\text{\textbf{Conventional Update: }       }
    W-\eta G = W - \eta \sum\limits_{i,j}g_{ij} u_iv_j^T\hspace{4.2cm}
    \label{eq:update1v1}
\end{equation}
\vspace{0cm}
\begin{equation}
\text{\textbf{MLET Update: }}
    W - \eta (W_1W_1^TG+ GW_2^TW_2) = W - \eta \sum\limits_{i,j}g_{ij}(\sigma_1(i)^2+\sigma_2(j)^2) u_iv_j^T
    \label{eq:update3v1}
\end{equation}
\begin{flushleft}
\vspace{0cm}
\textbf{Proof.} Eq.\ref{eq:update1v1} follows from \textbf{claim 2}. To derive Eq.\ref{eq:update3v1},
recall that the SVDs of $W_1$ and $W_2$ are $W_1 = U\Sigma_1X^T$ and $W_2 = Y\Sigma_2V^T$.
Let $I_k$ denote the identity matrix of shape $k\times k$.
\end{flushleft}
\vspace*{-0.5cm}

\begin{equation}
\begin{split}
    &\mathrm{vec}(W_1W_1^TG + GW_2^TW_2) \\
    = & \text{ vec}(U\Sigma_1\Sigma_1^TU^TGI_n) + \mathrm{vec}(I_dGV\Sigma_2^T\Sigma_2V^T)\\
    \overset{(a)}{=} &(I_n\otimes U\Sigma_1\Sigma_1^TU^T)\mathrm{vec}(G)+ (V\Sigma_2^T\Sigma_2V^T\otimes I_d)\mathrm{vec}(G)\\
    \overset{(b)}{=}&\left((VI_nV^T\otimes U\Sigma_1\Sigma_1^TU^T)+ (V\Sigma_2^T\Sigma_2V^T\otimes UI_dU^T)\right)\mathrm{vec}(G)\\
    \overset{(c)}{=}&\left((V\otimes U)(I_n\otimes \Sigma_1\Sigma_1^T)(V^T\otimes U^T) + (V\otimes U)(\Sigma_2^T\Sigma_2\otimes I_d)(V^T\otimes U^T)\right)\mathrm{vec}(G)\\
    \overset{(d)}{=}&\left((V\otimes U)(I_n\otimes \Sigma_1\Sigma_1^T+\Sigma_2^T\Sigma_2\otimes I_d)(V\otimes U)^T\right)\mathrm{vec}(G)\\
    =&\sum\limits_{i,j}(v_j\otimes u_i)(\sigma_{1}(i)^2+\sigma_{2}(j)^2)(v_j\otimes u_i)^T\sum\limits_{i,j}g_{ij}(v_j\otimes u_i)\\
    \overset{(e)}{=}&\sum\limits_{i,j}g_{ij}(\sigma_{1}(i)^2+\sigma_{2}(j)^2)(v_j\otimes u_i)
    \label{apdx:4}
\end{split}
\end{equation}

In Eq.\ref{apdx:4}, (a) uses the property $\mathrm{vec}(ABC)=(C^T\otimes A)\mathrm{vec}(B)$.
(b) follows from the fact that $V$ and $U$ are orthogonal matrices.
(c) uses the property $ABC\otimes DEF=(A\otimes D)(B\otimes E)(C\otimes F)$.
(d) uses the property $(A\otimes B)^T=(A^T\otimes B^T)$.
(e) simplifies the equation by using the fact that $(v_j\otimes u_i)$s are orthornormal.
Notice that $(v_j\otimes u_i)$ is the vectorization of $(u_i v_j^T)$.
Therefore, when converted into matrix form, Eq.\ref{apdx:4} is equivalent to Eq.\ref{eq:update3v1}.
With that we conclude our proof.\hfill $\square$


\vspace{0cm}
\begin{figure}[H]
	\centering

	\includegraphics[width=13cm]{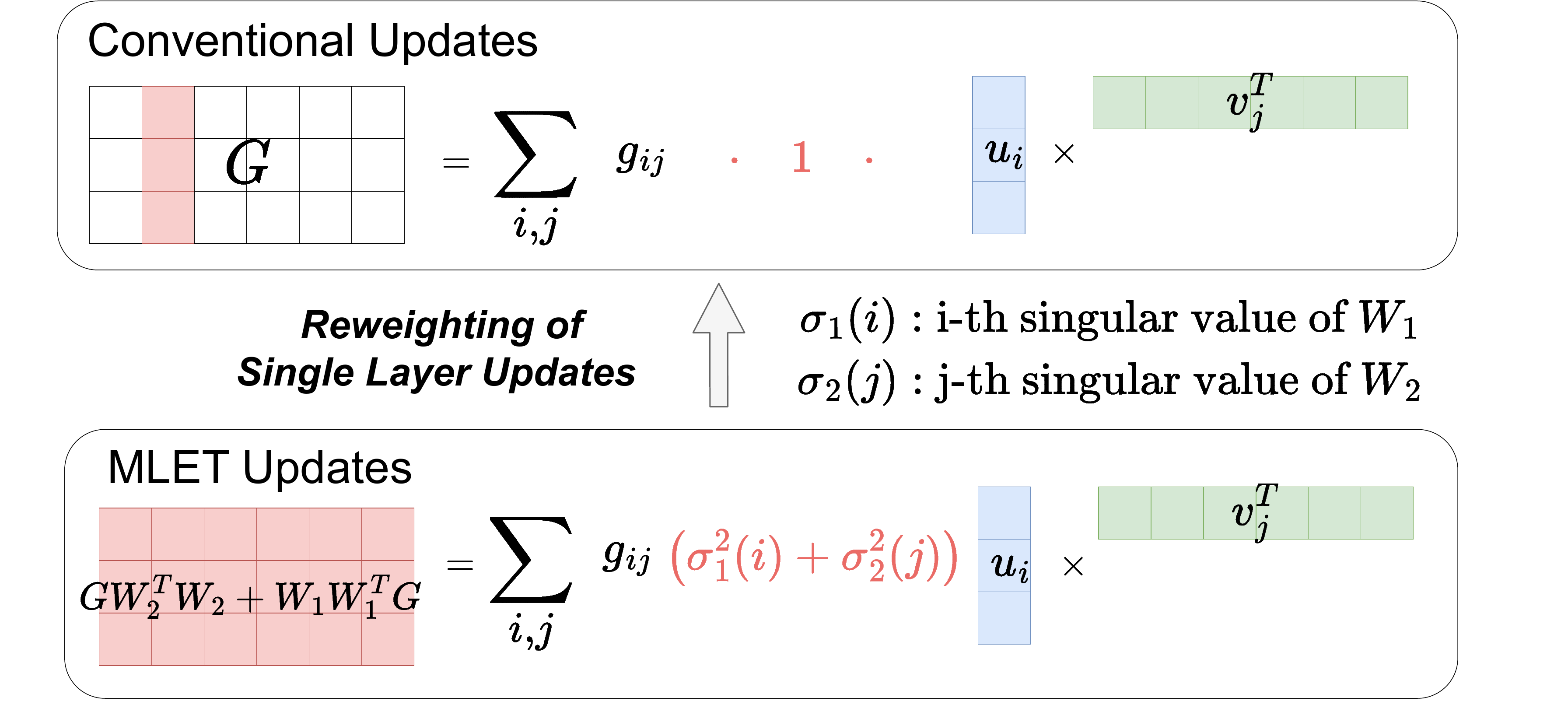}
    \caption{The reweighting mechanism of MLET: history-aware adjustments are based on the importance of update directions to learned embeddings, which is indicated by the singular values of MLET embeddings.} 
    \vspace{-0.6cm}
	\label{fig:theory}
\end{figure}
\vspace{0cm}

This theorem re-formulates the embedding updates as weighted sums of a set of base matrices, formed by outer products of singular vectors of embeddings.
\emph{It pins down the source of cross-category information to a reweighting process guided by embeddings singular values.
}
{\color{black}
The reweighting operates as follows (Fig.~\ref{fig:theory}).
In every training iteration, it reweights the embedding updates in each update direction $u_iv_j^T$ by a factor $(\sigma_1(i)^2+\sigma_2(j)^2)$.
Generally, a large singular value implies that the associated singular vector captures a significant amount of the structure or information within the data.
In MLET's reweighting mechanism, $\sigma_1(i)$ and $\sigma_2(j)$ indicate the importance of their associated singular vectors, $u_i$ and $v_j$, to the learned embeddings $W_1$ and $W_2$.
In this way, the reweighting factor boosts the update in the direction which has proven to be important based on earlier training history.

Such reweighting creates a similar effect to that of momentum in gradient descent. 
Momentum \cite{Sutskever2013OnTI,Ruder2016AnOO} adds a fraction of history updates to new updates.
It has been shown to mitigate oscillations and overshooting in the optimization process, and allow the algorithm to ``roll'' faster on shallow regions and navigate more effectively through complex loss landscapes.
\color{black}Unlike momentum based methods, MLET does not explicitly calculate the exponential moving average of the current and previous gradients. Rather, it implicitly achieves a similar effect by using the information in the learned embeddings (which is ignored by the standard momentum methods). It is evident that the embeddings themselves are the best capture of the past gradient updates and the long-term trend of the update directions. Reweighting reinforces the update along important directions of embeddings by providing positive feedback.
}

{\color{black}
\subsection{{\color{black}{Effect of Inner Dimension}}} 

\label{section-k}
The inner dimension $k$ is important.
Empirically, MLET requires $k>d$ to achieve superior performance and higher $k$s consistently achieve higher performance. 
Since the inference-time embedding table is of size $n\times d$, MLET with $k>d$ introduces more parameters ($nk+kd$) than needed ($nd$) and overparameterizes the model. \emph{Why does MLET require $k>d$ to make consistent improvement and why do the benefits of such overparameterization increase with $k$?}

The number of informative reweighting factors help answer the above two questions.
We note that in single-layer training (Eq.\ref{eq:update1v1}), the reweighting factor of all update directions $u_iv_j^T$ can be treated as constant 1.
One can show that the number of reweighting factors with a non-zero $\sigma_2$ is $kd$ for MLET with inner dimension $k$.
$\sigma_2(j)$ measures the importance of $v_j$ to the embedding table so it is informative in determining the confidence in taking update $u_i^Tv_j,i\in\{1,2,..,d\}$.
Intuitively, if $\sigma_2=0$, the informativeness of reweighting is reduced.

For $k\ge d$, consider two MLET models with inner dimensions $k_\mathit{big},k_\mathit{small}$ ($k_\mathit{big}>k_\mathit{small}\ge d$).
The model with inner dimension $k_\mathit{small}$ has $d(k_\mathit{big}-k_\mathit{small})$ fewer informative reweighting factors because of $\sigma_2=0$.
Being less informative in updates generally leads to worse training performance of $k_{small}$.

For $k<d$, not only the number of informative factors reduces with smaller $k$, the number of inactive(zero) factors increases.
In this case, there are $(n+d-k)k$ non-zero reweighting factors in MLET. 
(To see this, consider that the number of factors with at least one of $\sigma_1(i)$ and $\sigma_2(j)$ being non-zero.
For $i\in\{1,..,k\}$, all $\sigma_1(i)$s are non-zero, so their related reweighting factors are non-zero and there are $k\times n$ such factors.
For $i\in\{k+1,..,d\}$, all $\sigma_1(i)$s are zero and reweighting factors are non-zero only when $j\in \{1,..,k\}$.
There are $(d-k)k$ such factors. Thus, there are $kn+(d-k)k$ non-zero enhancement factors in total.)
However, the number of non-zero reweighting factors in the single-layer training is $dn$.
Because $dn-(n+d-k)k=(n-k)(d-k)>0$, single-layer training has $(n-k)(d-k)$ more flexible update directions that cannot be taken by MLET (because MLET assigns zero reweighting factors to them).
For such MLET models, this lack of flexibility in training updates worsens their performance.

\vspace{0cm}
\begin{table}[H]

    \centering
    
    \begin{tabular}{ccccccccccc}
        \hline
         \multirow{2}{*}{Update Direction}& \multirow{2}{*}{$u_1v_1^T$} & \multirow{2}{*}{$u_1v_2^T$} & \multirow{2}{*}{$u_1v_3^T$} & \multirow{2}{*}{$u_1v_4^T$} & \multirow{2}{*}{$u_1v_5^T$} & \multirow{2}{*}{$u_2v_1^T$} & \multirow{2}{*}{$u_2v_2^T$} & \multirow{2}{*}{$u_2v_3^T$} & \multirow{2}{*}{$u_2v_4^T$} & \multirow{2}{*}{$u_2v_5^T$} \\
         &\\
         \hline
         Single-Layer & 1 & 1 & 1 & 1 & 1 & 1 & 1 & 1 & 1 &1 \\
         MLET (k=1) & {\color{red}$\checkmark$} & $\checkmark$ & $\checkmark$ & $\checkmark$ & $\checkmark$ & 0 & 0 & 0 & 0 & 0\\
         MLET (k=2) & {\color{red}$\checkmark$} & {\color{red}$\checkmark$} & $\checkmark$ & $\checkmark$ & $\checkmark$ & {\color{red}$\checkmark$} & {\color{red}$\checkmark$} & $\checkmark$ & $\checkmark$ & $\checkmark$\\
         MLET (k=4) & {\color{red}$\checkmark$} & {\color{red}$\checkmark$} & {\color{red}$\checkmark$} & {\color{red}$\checkmark$} & $\checkmark$ & {\color{red}$\checkmark$} & {\color{red}$\checkmark$} & {\color{red}$\checkmark$} & {\color{red}$\checkmark$} & $\checkmark$ \\
         \hline
         \vspace{0cm}
    \end{tabular}
    \caption{Reweighting factors for embeddings with $(d=2,n=5)$. $\checkmark$: active factors informative of $W_1$ (with non-zero $\sigma_1$). {\color{red}$\checkmark$}: active factors informative of both $W_1$ and $W_2$ (with non-zero $\sigma_1$ and $\sigma_2$). }
    \label{tab:innerDim}
\end{table}
\vspace{0cm}

To illustrate the above points, Table~\ref{tab:innerDim} presents, for a toy case with small dimensions, the reweighting factors for different training schemes.
In $k<d$, half of the MLET reweighting factors are inactive/zero. 
A larger $k$ leads to more active and more informative factors.
}

\section{Experiments}
\vspace{0cm}
We evaluate the proposed MLET technique on seven state-of-the-art recommendation models on two public datasets for click-through rate tasks: Criteo-Kaggle \cite{CriteoKaggle2014} and Avazu \cite{AvazuKaggle2014}.
Both datasets are composed of a mix of categorical and real-valued features (Table~\ref{tab:dataset_info}). 
The Criteo-Kaggle dataset was split based on the time of data collection: the first six days are used for training and the seventh day is split evenly into the test and validation sets. 
The Avazu dataset was randomly split into training and test sets of 90\% and 10\%, respectively.
The models are implemented in PyTorch and trained on systems with NVIDIA GPUs (CUDA acceleration enabled).

\vspace{0cm}
\begin{table}
\centering
\caption{Dataset composition.}
\begin{tabular}{@{}llll@{}}
\toprule
Dataset       & Samples & Dense Features & Sparse Features \\ \midrule
Criteo-Kaggle \cite{CriteoKaggle2014} & 45,840,617    & 13             & 26                   \\
Avazu \cite{AvazuKaggle2014}        & 40,400,000    & 1              & 21                   \\ \bottomrule
\end{tabular}
\label{tab:dataset_info}
\end{table}
\vspace{0cm}
Seven state-of-the-art recommendation models are evaluated. 
DLRM is tested both on Criteo-Kaggle and Avazu. 
Other models are tested exclusively on the Avazu dataset because of its reduced runtime requirement relative to Criteo-Kaggle. We use publicly available implementations of non-DLRM models from the open-source recommendation model library DeepCTR-Torch \cite{shen2019deepctrtorch}.
To decrease the impact of randomized initialization and run-to-run variation due to non-deterministic GPU execution, the reported results are averaged using at least three training runs. We report two quality metrics: area under the ROC curve (AUC) and binary cross-entropy (LogLoss).
\vspace{0cm}
\begin{figure}
\centering
\includegraphics[width=12cm]{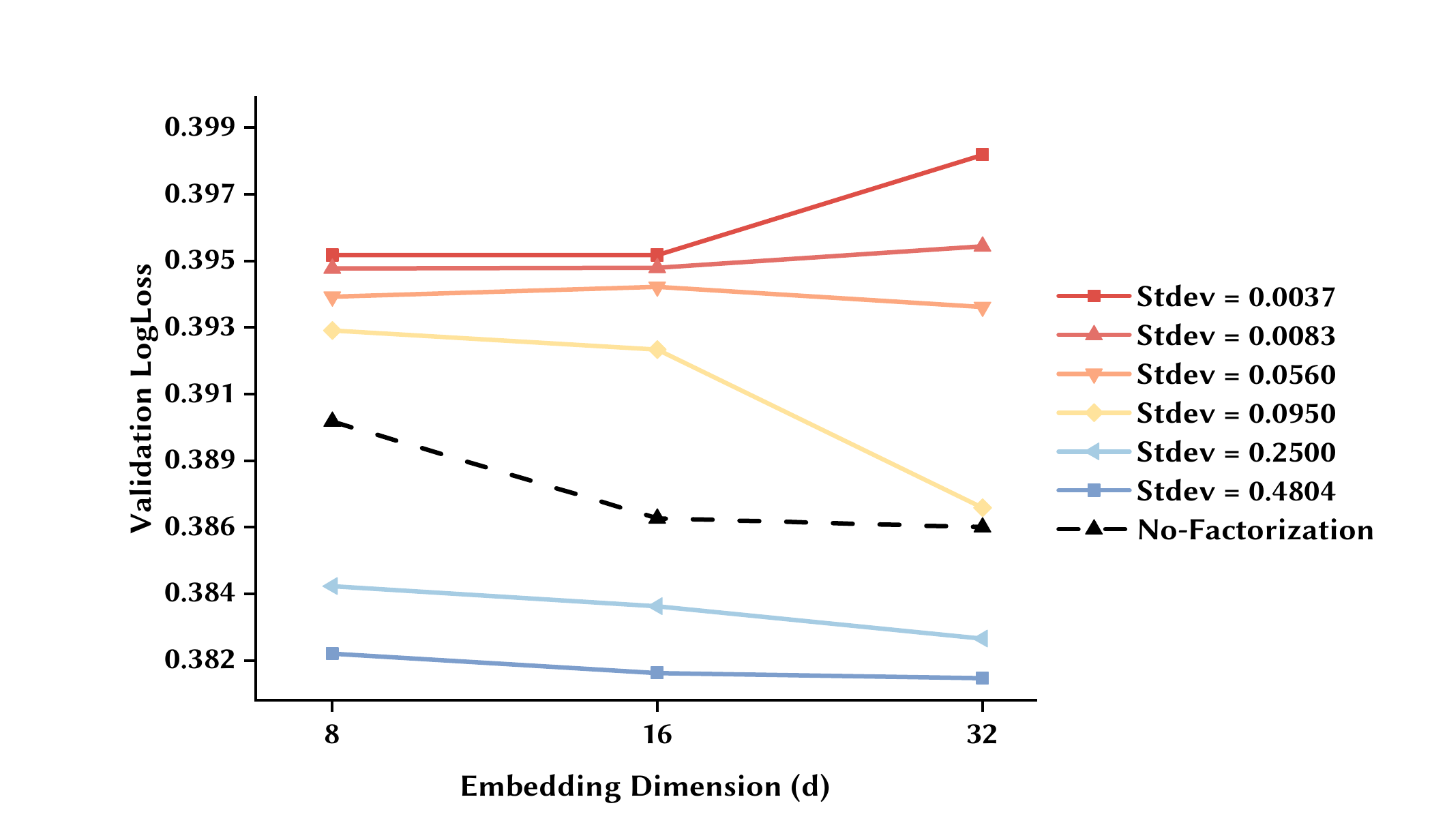}
\vspace*{-0.5cm}
\caption{Search of initialization variance (embedding factorization) for DLRM  with MLET $k=32$.
An appropriately large variance is critical to MLET's effectiveness. Small variance leads to vanishing reweighting factors hence undertrained performance.}
\vspace*{-1cm}
\label{fig:variance}
\end{figure}

Initialization strategy used for embedding layers is of critical importance in training RMs.
In conventional RMs, the embedding table of each sparse feature is represented by a single linear layer. 
We follow a conventional approach in initializing this layer that uses Xavier initialization scheme \cite{Glorot2010UnderstandingTD}. 
MLET adds another linear factorization layer. We use a Gaussian distribution to initialize this second factorization layer.
To make MLET effective, initialization variance cannot be too small.
As suggested by \textbf{Theorem 1}, small initialization effectively leads to vanishing reweighting factors and slows down embedding updates.
This results in poor performance as shown in Figure \ref{fig:variance}.
Empirically, if variance is too high then the training suffers from convergence issue.
In all the following experiments, we set the initialization standard deviation to 0.25 for DLRM and 0.5 for other models unless otherwise noted.
Those values ensure the effectiveness of MLET while preserving training-time convergence.

Following prior work~\cite{Naumov2019DeepLR}, we train all models for a single epoch {\color{black}to avoid over-fitting}.
Two optimizers are tested: SGD and Adagrad. 
DLRM and its MLET variants are trained using SGD with a learning rate of $0.2$.
Other models are trained using Adagrad with a learning rate of $0.02$.
{\color{black}The above learning rates achieve the optimal/near-optimal conventional single-layer embedding training results and the improvements that are possible by changing them are negligible.}
In all experiments, $d$ stands for embedding dimension. 
For DLRM, on both datasets we configure its top MLP to have two hidden layers with 512 and 256 nodes. 
On the Avazu dataset, we set DLRM's bottom MLP to be $256 \rightarrow 128 \rightarrow d$. 
On the Criteo-Kaggle dataset, we configure DLRM’s bottom MLP to be $512 \rightarrow 256 \rightarrow 128 \rightarrow d$. 
{\color{black}
Other models use the default model architectures and hyperparameters from the DeepCTR library. }
\vspace{0cm}
\subsection{Learning Enhancement}
\vspace{0cm}
\label{section-experiment}
The experiments demonstrate the effectiveness of MLET in producing superior models compared to the baseline single-layer embedding training. Figures \ref{fig:edim_vs_logloss} and \ref{fig:edim_vs_auc} summarize the experiments with DLRM carried out on two datasets.
Figure \ref{fig:alternate-models} presents the main results for three other models: DCN, NFM, and AutoInt. 
Table~\ref{tab:summary_multi_model} summarizes the results of MLET on all the seven models we tested.
{\color{black}
The maximum memory reduction is calculated using all the data points with different $k,d$ combinations we tested (similar to Figure \ref{fig:alternate-models}).

As Figures \ref{fig:edim_vs_logloss} and \ref{fig:edim_vs_auc} show, MLET consistently squeezes more performance out of fixed-size embeddings of DLRM model.
The benefits begin to be observed in MLET curves even for $k=d$. Increasing $k$ for a given $d$ leads to a monotonic improvement in model accuracy. 
\textit{For CTR systems, an improvement of $0.1\%$ in AUC is considered substantial}.
The maximum AUC benefit of MLET for Criteo-Kaggle is $0.27\%$, and the maximum benefit for Avazu is $1.24\%$. This improvement in model accuracy saturates as $k$ grows, e.g., on the Criteo-Kaggle dataset the curves with $k=64$ and $k=128$ are very similar. 

As can be seen in Figures \ref{fig:edim_vs_logloss} and \ref{fig:alternate-models}, the general performance vs.\ vector dimension behavior is similar across the different models evaluated. We note that not only is the overall behavior similar, but also that MLET provides substantial benefits for most models with $4-16\times$ savings of embedding parameters while maintaining the same or better performance as compared to the single-layer embedding training.

\begin{figure}[H]

	\subfigure{
		\includegraphics[width=7cm,height=5cm]{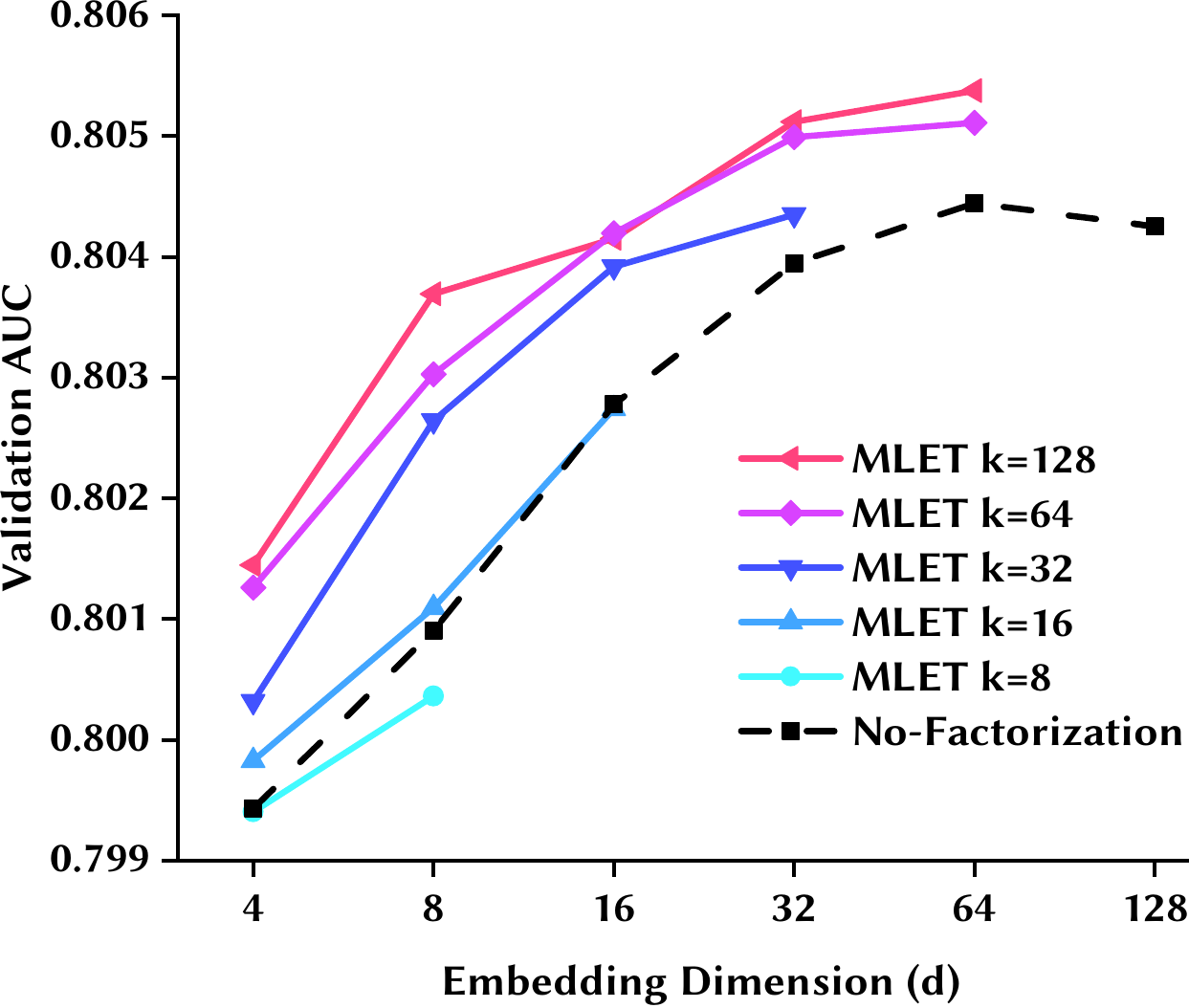}}
	\subfigure{
		\includegraphics[width=7cm,height=5cm]{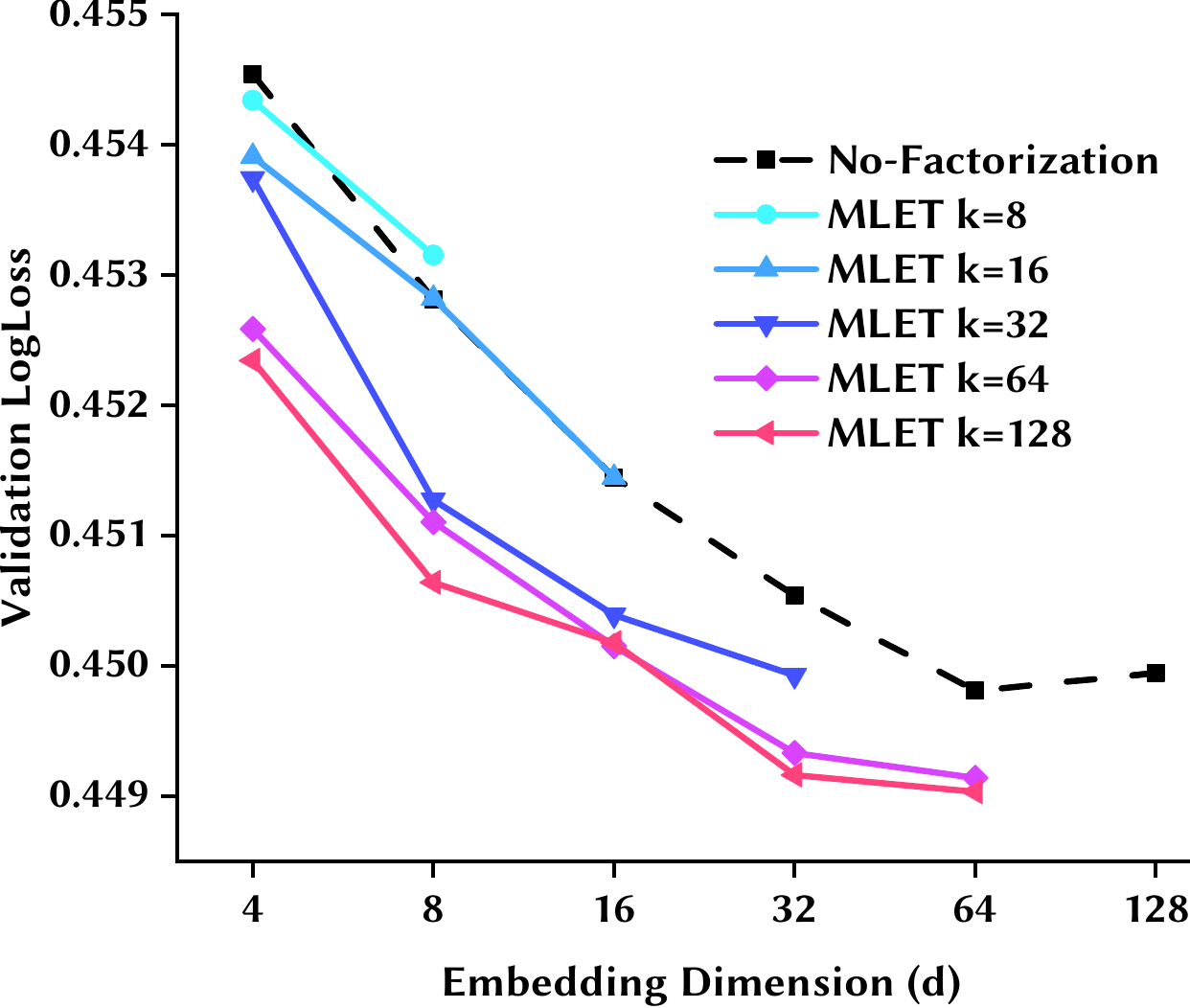}}

	\caption{MLET with DLRM on the Criteo-Kaggle dataset.}
        \vspace*{-0.5cm}
	\label{fig:edim_vs_logloss}
\end{figure}

\vspace{0cm}
\begin{figure}[H]	

    \subfigure{
		\includegraphics[width=7cm,height=5cm]{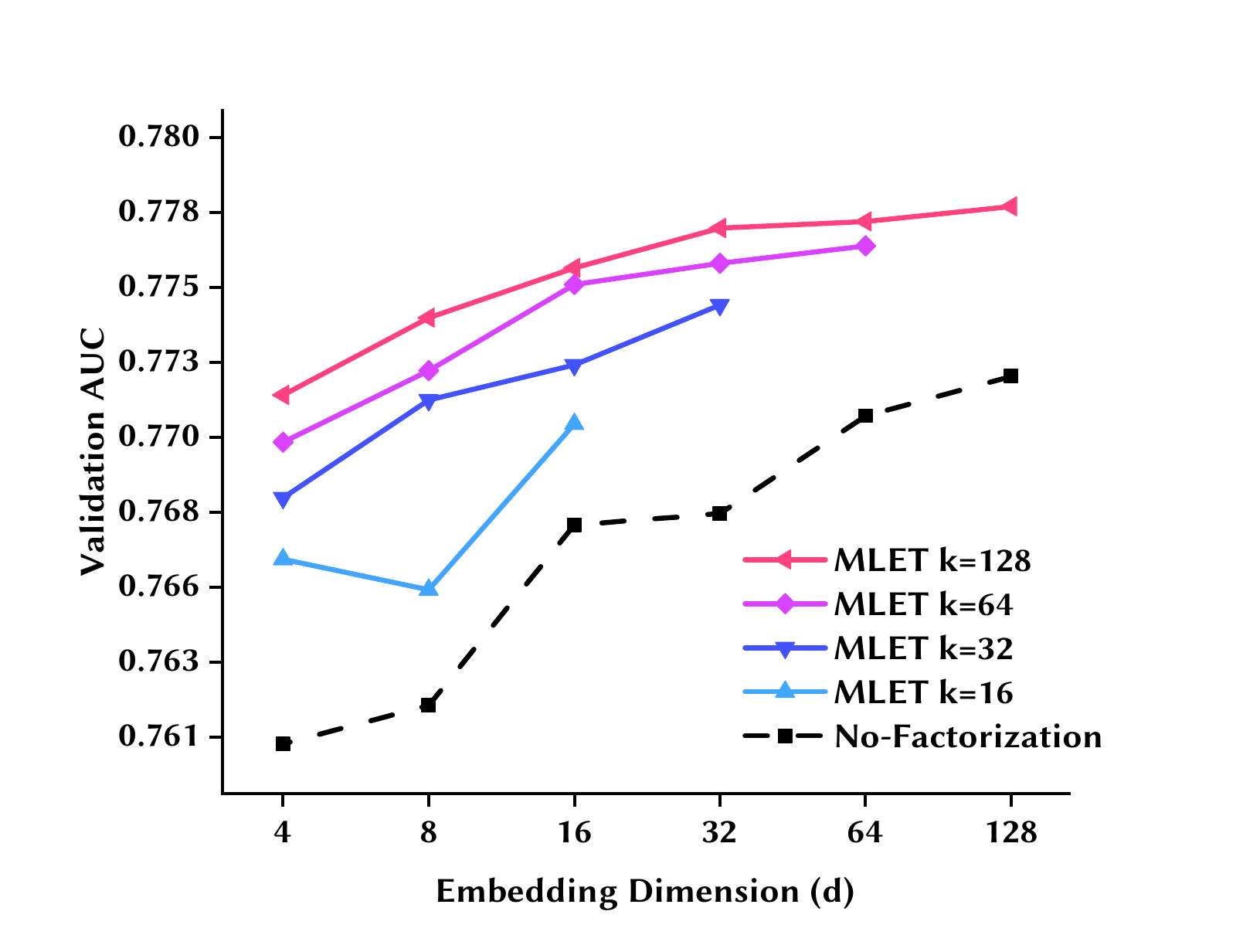}}
    \subfigure{
        \includegraphics[width=7cm,height=5cm]{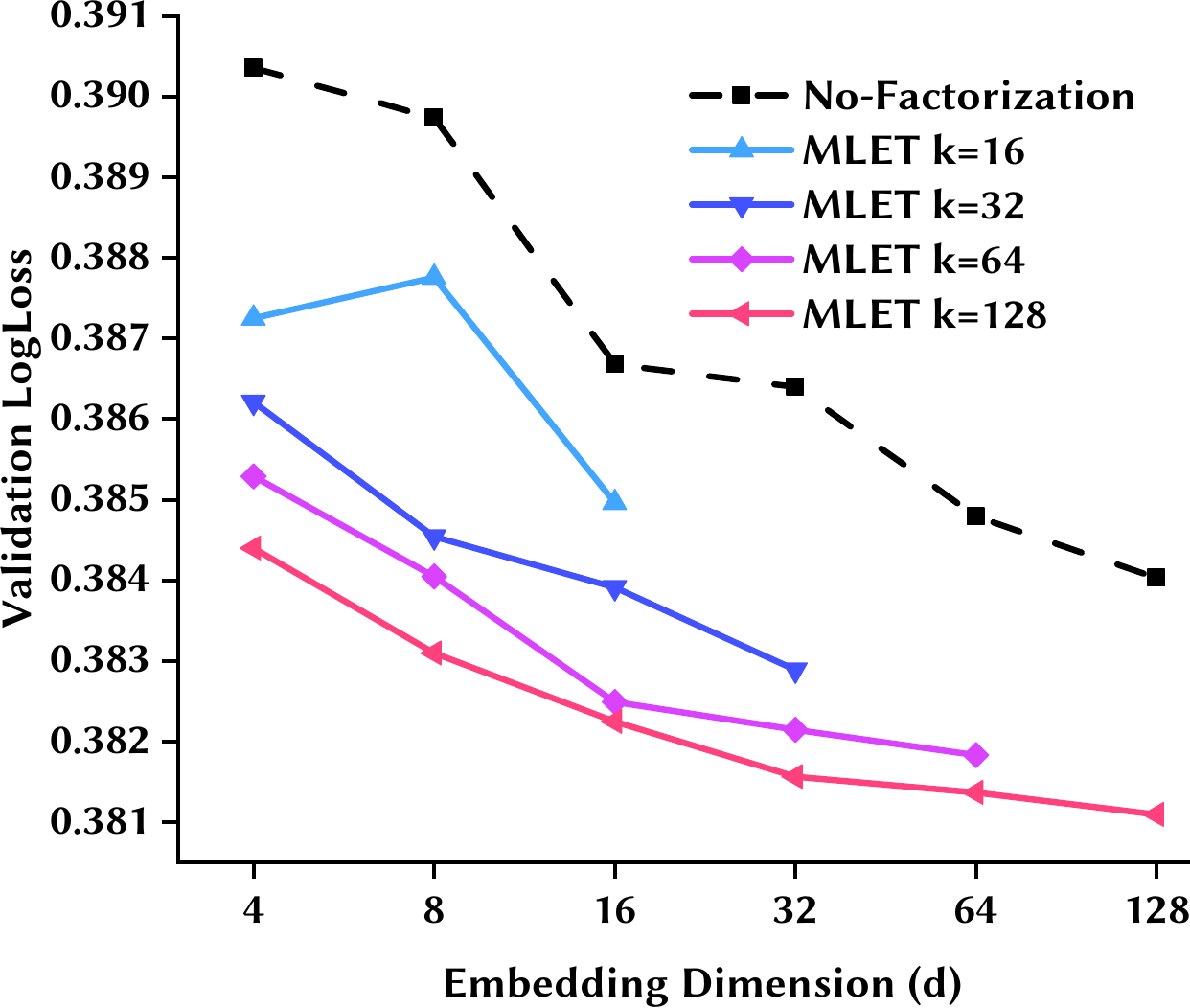}}
    \vspace{0cm}
    \caption{MLET with DLRM on the Avazu dataset.}
    \label{fig:edim_vs_auc}
\end{figure}

\begin{figure}[H]
	\subfigure[AutoInt]{
		\includegraphics[width=5cm]{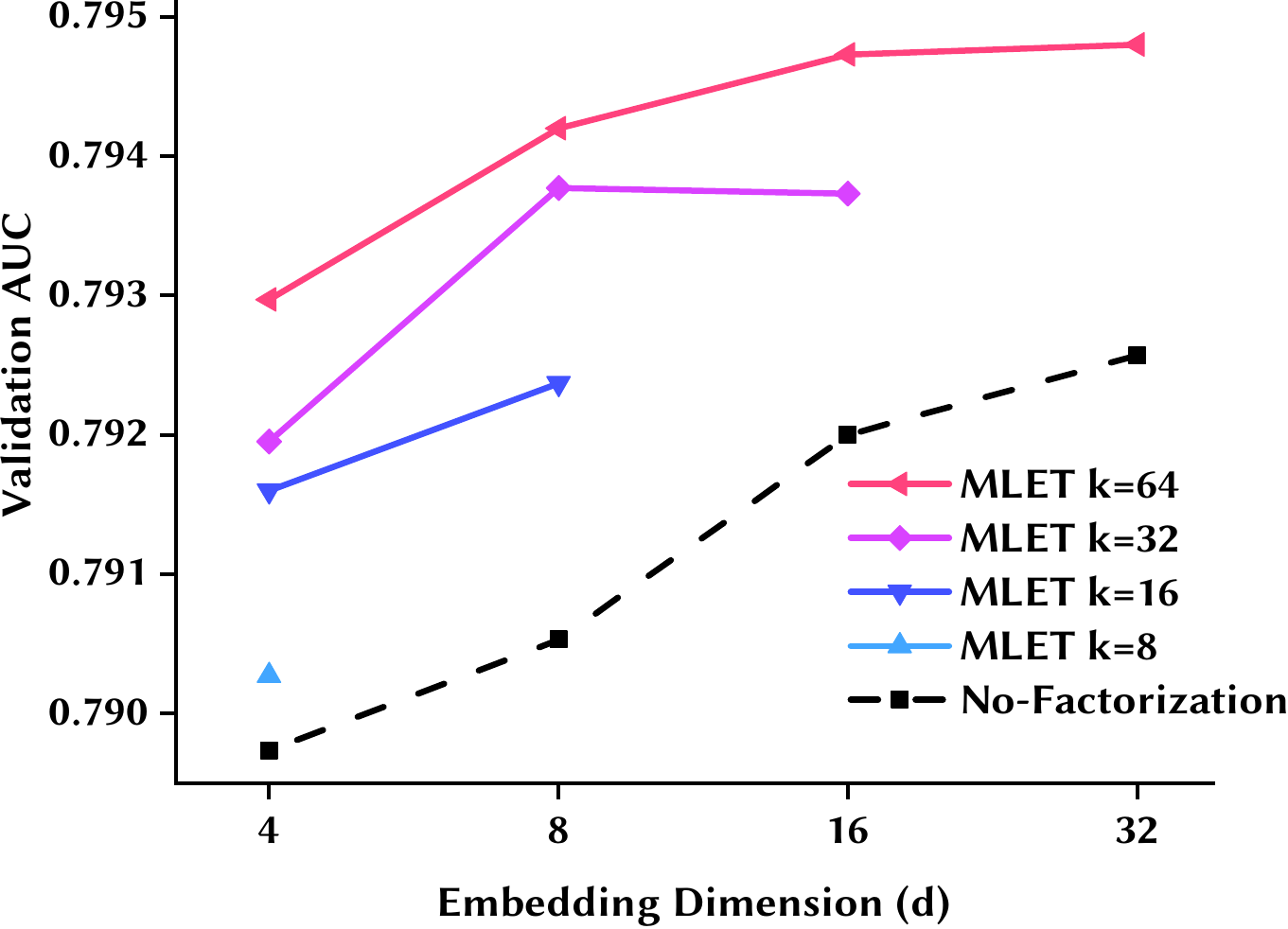}}
	\subfigure[DCN]{
		\includegraphics[width=5cm]{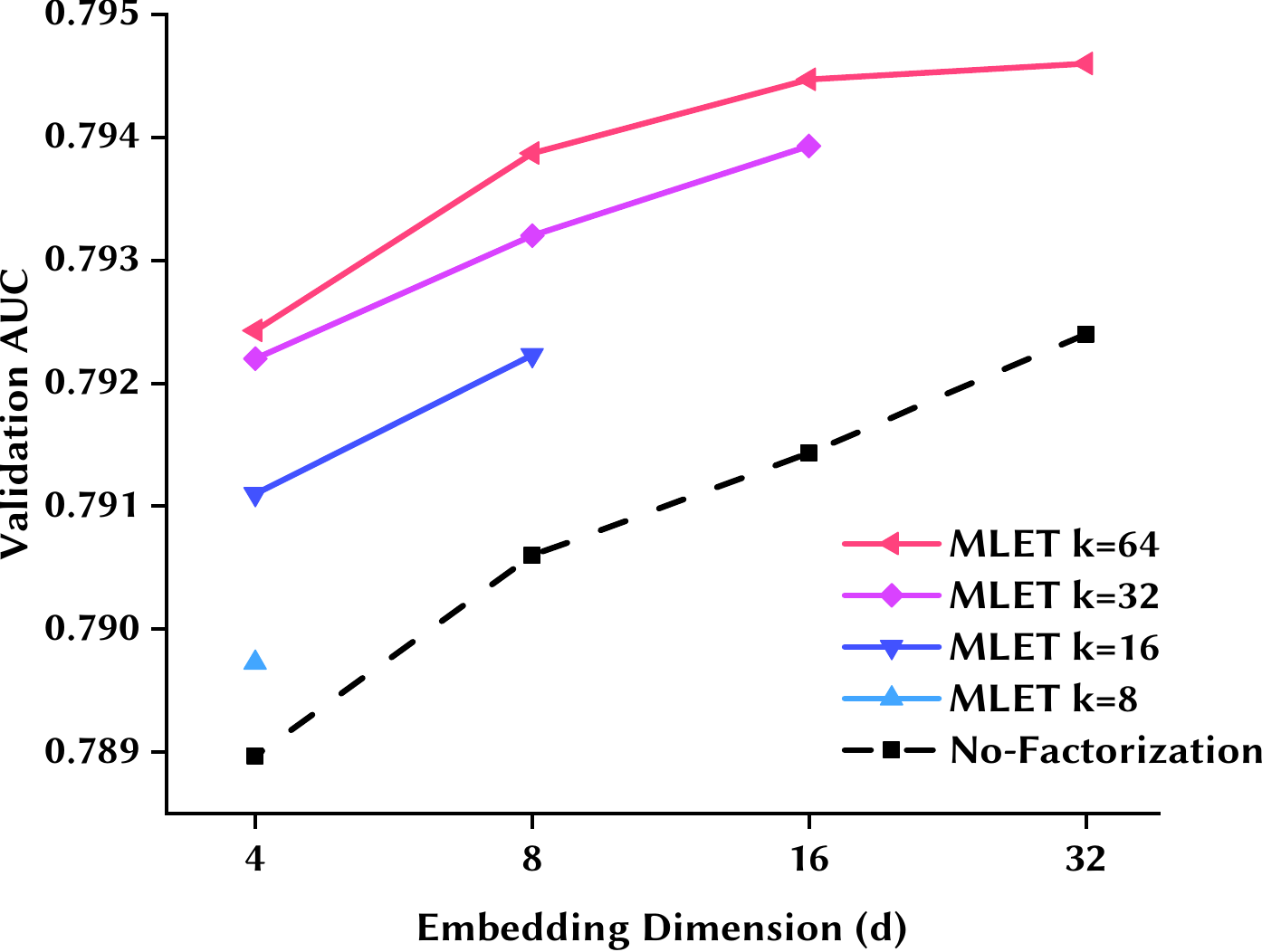}}
	\subfigure[NFM]{        \includegraphics[width=4.5cm]{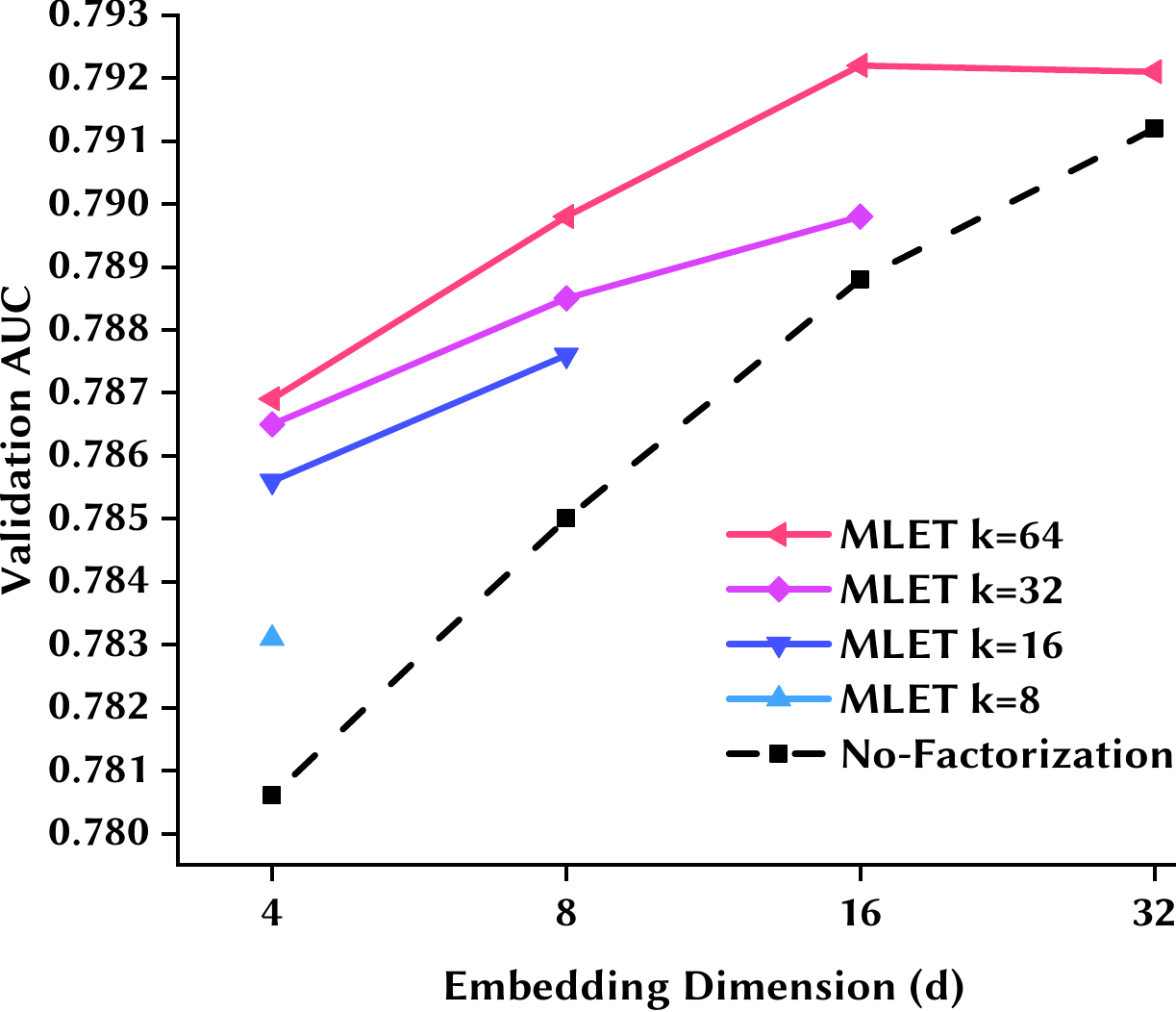}	
	}
        \vspace*{-0.5cm}
	\caption{MLET on several state-of-the-art RM models on the Avazu dataset.}
        \vspace*{-0.5cm}
	\label{fig:alternate-models}
\end{figure}

\vspace{0cm}
\begin{table}[H]
\small
\begin{tabular}{@{}llccccl@{}}
\toprule
 & & \multicolumn{2}{c}{Baseline AUC} & \multicolumn{2}{c}{MLET AUC ($k,d$)}& Maximum Memory Reduction\\
 Model         & Dataset & $d=4$ & $d=16$ &  64,4 &  64,16 & (same or higher performance)\\ \midrule
 
 \multirow{2}{*}{DLRM} & Criteo  & 0.799 & 0.803 & 0.801 & 0.804 & 4x ($d\,32\rightarrow8\,,k\ge 128$)\\
 & Avazu  &                           0.761 & 0.768 & 0.770 & 0.775 & 16x($d\,64\rightarrow4\,,128\rightarrow8\,,k\ge 128$)\\
WDL  &Avazu  &        0.789 & 0.790 & 0.792 & 0.793 & 4x ($d\,16\rightarrow4\,,k\ge 64$)\\
DeepFM        &Avazu  &         0.790 & 0.794 & 0.792 & 0.795 & 1x\\
xDeepFM       & Avazu   &   0.792 & 0.796 & 0.794 & 0.798 & 2x ($d\,16\rightarrow8\,,k\ge 64$)\\
DCN &Avazu           & 0.789 & 0.791 & 0.792 & 0.794 & 8x ($d\,32\rightarrow8\,,16\rightarrow4\,,k\ge 64$)\\
AutoInt       &Avazu &           0.790 & 0.792 & 0.793 & 0.795 & 8x ($d\,32\rightarrow4\,,k\ge 32$)\\
NFM           &Avazu &     0.781 & 0.789 & 0.787 & 0.792 & 4x ($d\,16\rightarrow4\,,k\ge 64$)\\
 
 \bottomrule
\end{tabular}
\caption{Effectiveness of MLET.}
\label{tab:summary_multi_model}
\end{table}

\subsection{Learning Quality for High- and Low-Frequency Embeddings}
Since embedding updates of MLET are cross-category informative and are more frequent, they should lead to better learning quality of embeddings, especially those of the least frequently queried categories.
To verify this intuition, 
we conduct experiments that compare the performance of MLET and that of single-layer training on two test sets.
Set (A) is composed by $10\%$ test samples with the most frequently queried categories.
Set (B) is composed by $10\%$ test samples with the least frequently queried categories.
{\color{black}
To sort the samples, we first calculate on the training set the frequencies of all categories in each sparse feature.
Then the frequency of each test sample is estimated by multiplying the frequencies of all categories it queries.
}
Experiments are done with three models (DCN, AutoInt, and xDeepFM) on the Avazu dataset.
We use the relative improvement in PR-AUC to evaluate MLET's enhancement in the learning quality of embeddings.
Since 20\% of A are clicked while only 15\% of B are clicked,
we use PR-AUC instead of ROC-AUC because it is more robust to imbalanced data and is more sensitive to the improvements for the positive class. \cite{AUCCompares}.

As shown in Table~\ref{tab:embd_quality}, MLET generally improves embedding quality on both sets of samples.
Further, MLET consistently improves performance on the least frequent samples (set B) and the improvements on them are larger than the improvements on the most frequent samples (set A). 
This empirical observation aligns with our expectation from the theory that MLET's dense and cross-category informative updates are most beneficial to the learning quality of the embeddings of rarely-occurring categories.

\subsection{Combining MLET with Post Training Model Compression}
\vspace{0cm}
{
We conduct experiments to test the comparison and composition of MLET with several commonly used post-training model compression techniques.

\justify{
\textbf{Low Rank SVD Approximation} As pointed out by \cite{bhavana2019block}, the numerical rank of embedding tables can be much smaller than their embedding dimension, and hence, SVD factorization allows the original embedding table to be stored and recovered inexpensively with the low-dimensional factor matrices. 
Table~\ref{tab:mlet_vs_svd} shows a comparison of MLET and a low-rank SVD approximation on three models at different embedding sizes.
For MLET, the embedding size is its embedding dimension.
For an SVD-compressed model, it is the number of reserved ranks in the low-rank approximation of its embedding tables, trained by conventional single-layer training.
For example, an SVD model with embedding size 16 means that the embedding tables are  approximated by rank-16 approximations.
We see that MLET maintains its advantage over SVD at embedding sizes. 
}
\begin{table}[H]

    \centering
    
    \begin{tabular}{ccc}
        \hline
         \multirow{2}{*}{Model(d/k)} & Set A  & Set B \\
          & (Most Frequent) & (Least Frequent) \\
         \hline
         DLRM(16/64) & +0.22\% & +1.08\%\\
         DCN(4/8) & +0.04\% & +0.13\% \\
         DCN(16/64) & +0.33\% & +0.45\% \\
         AutoInt(4/8) & +0.05\% & +0.08\% \\
         AutoInt(16/64) & +0.38\% & +0.48\% \\
         xDeepFM(4/8) & -0.09\% & +0.19\% \\
         xDeepFM(16/64) & +0.01\% & +0.24\% \\
         \hline
         \vspace*{-0.5cm}
    \end{tabular}
    \caption{Improvement of PR-AUC on test samples with the most/least frequently queried items: low-frequency embeddings benefits more from MLET.}
    \vspace*{-0.5cm}
    \label{tab:embd_quality}
\end{table}

\vspace*{-0.5cm}
\begin{table}[H]

    \centering
    
    \begin{tabular}{cccccc}
        \hline
         \multirow{2}{*}{Model} & \multirow{2}{*}{Configuration} &\multicolumn{4}{c}{Effective Embedding Size} \\
          & & 4 & 8 & 16 & 32 \\
         \hline
         \multirow{4}{*}{DCN} & SVD d=32 & 0.7783 & 0.7903 & 0.7914 & - \\
         & MLET k=32 & 0.7922 & 0.7932 & 0.7939 & - \\
         & SVD d=64 & 0.7659 & 0.7902 & 0.7923 & 0.7927 \\
         & MLET k=64 & 0.7924 & 0.7939 & 0.7945 &0.7946 \\
         \hline 
         \multirow{4}{*}{AutoInt} & SVD d=32 &0.7812  &0.7916  &0.7929 & - \\
         & MLET k=32 &0.7920 &0.7937  &0.7938  & - \\
         & SVD d=64 &0.7761 & 0.7910  &0.7927& 0.7930\\
         & MLET k=64 &0.7930 & 0.7942&0.7947&0.7948 \\
         \hline 
         \multirow{4}{*}{xDeepFM} & SVD d=32 &0.7672  &0.7818 &0.7920  & - \\
         & MLET k=32 &0.7933 & 0.7955  & 0.7972  & - \\
         & SVD d=64 &0.7618 & 0.7783  & 0.7895 & 0.7962 \\
         & MLET k=64 &0.7935&0.7957 &0.7978 & - \\
         \hline 
         \vspace*{-0.5cm}
    \end{tabular}
    \caption{MLET vs. Low Rank SVD Approximation of embedding tables: with the same size of embeddings (at both training and inference), MLET produces better models.  }
    \vspace*{-1cm}
    \label{tab:mlet_vs_svd}
\end{table}
\vspace{0cm}

\vspace{0cm}
\justify{
\textbf{Quantization and Hashing}
We use quantization on the trained model \cite{Krishnamoorthi2018QuantizingDC}, quantizing the embedding tables to 8 bits. We leave the rest of the model in full precision.
A uniform symmetric quantizer is used, with its scaling factors determined via a grid search that minimizes the L2 error between the FP32 embeddings and their quantized values.
The hashing trick, as described in \cite{attenberg2009collaborative}, reduces table width by hashing the indices of categories into a smaller index space. We use the modulo hash function to hash the two largest tables in the Avazu to half of their original sizes. These two tables (device\_ip and device\_id) jointly account for 99.7\% of all embeddings. We do not hash other tables, as the resulting model size savings are negligible. 
Figure \ref{fig:composition_of_techniques} presents the results of experiments performed on the DCN model.
MLET improves model quality with all combinations of quantization and hashing. 
}
\begin{figure}[H]
\centering
    \includegraphics[width=14cm]{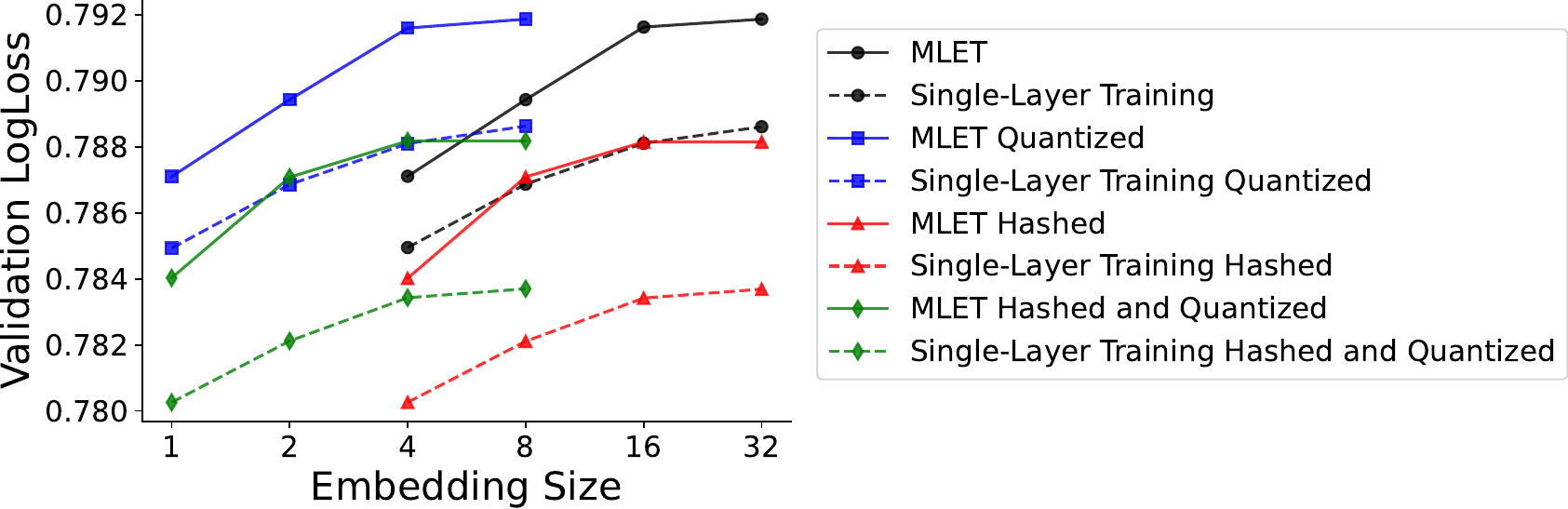}
\caption{Composition of MLET with quantization and hashing. 
}
\vspace*{-0.8cm}
\label{fig:composition_of_techniques}
\end{figure}

\vspace{0cm}
\section{Conclusion}
\vspace{0cm}

{\color{black}We introduce a strikingly simple yet effective multi-layer embedding training (MLET) architecture that trains embeddings via a sequence of linear layers to derive superior models. We present a theory that explains the superior embedding learning via the dynamics of embedding updates. We prototype MLET across seven state-of-the-art open-source recommendation models and demon- strate that MLET alone is able to achieve the same or better performance as compared to conventional single-layer training scheme while uses up to 16x less (5.8x less on average) embedding parameters.}
\vspace*{-0.4cm}
\bibliography{acml23}

\begin{thebibliography}{32}
\providecommand{\natexlab}[1]{#1}
\providecommand{\url}[1]{\texttt{#1}}
\expandafter\ifx\csname urlstyle\endcsname\relax
  \providecommand{\doi}[1]{doi: #1}\else
  \providecommand{\doi}{doi: \begingroup \urlstyle{rm}\Url}\fi

\bibitem[Alvarez and Salzmann(2017)]{alvarez2017compression}
Jose~M Alvarez and Mathieu Salzmann.
\newblock Compression-aware training of deep networks.
\newblock In \emph{Advances in Neural Information Processing Systems}, pages 856--867, 2017.

\bibitem[Arora et~al.(2018)Arora, Cohen, and Hazan]{Arora2018OnTO}
Sanjeev Arora, Nadav Cohen, and Elad Hazan.
\newblock On the optimization of deep networks: Implicit acceleration by overparameterization.
\newblock \emph{ArXiv}, abs/1802.06509, 2018.

\bibitem[Arora et~al.(2019)Arora, Cohen, Hu, and Luo]{Arora2019ImplicitRI}
Sanjeev Arora, Nadav Cohen, Wei Hu, and Yuping Luo.
\newblock Implicit regularization in deep matrix factorization.
\newblock In \emph{NeurIPS}, 2019.

\bibitem[Attenberg et~al.(2009)Attenberg, Weinberger, Dasgupta, Smola, and Zinkevich]{attenberg2009collaborative}
Josh Attenberg, Kilian Weinberger, Anirban Dasgupta, Alex Smola, and Martin Zinkevich.
\newblock Collaborative email-spam filtering with the hashing trick.
\newblock CEAS, 2009.

\bibitem[Bhavana et~al.(2019)Bhavana, Kumar, and Padmanabhan]{bhavana2019block}
Prasad Bhavana, Vikas Kumar, and Vineet Padmanabhan.
\newblock Block based singular value decomposition approach to matrix factorization for recommender systems.
\newblock \emph{arXiv preprint arXiv:1907.07410}, 2019.

\bibitem[Cheng et~al.(2016)Cheng, Koc, Harmsen, Shaked, Chandra, Aradhye, Anderson, Corrado, Chai, Ispir, Anil, Haque, Hong, Jain, Liu, and Shah]{WideDeep}
Heng{-}Tze Cheng, Levent Koc, Jeremiah Harmsen, Tal Shaked, Tushar Chandra, Hrishi Aradhye, Glen Anderson, Greg Corrado, Wei Chai, Mustafa Ispir, Rohan Anil, Zakaria Haque, Lichan Hong, Vihan Jain, Xiaobing Liu, and Hemal Shah.
\newblock Wide {\&} deep learning for recommender systems.
\newblock \emph{CoRR}, abs/1606.07792, 2016.
\newblock URL \url{http://arxiv.org/abs/1606.07792}.

\bibitem[Czakon(2021)]{AUCCompares}
Jakub Czakon.
\newblock F1 score vs roc auc vs accuracy vs pr auc: Which evaluation metric should you choose?
\newblock \url{https://neptune.ai/blog/f1-score-accuracy-roc-auc-pr-auc}, 2021.
\newblock [Online; Updated December 31st, 2021].

\bibitem[Ginart et~al.(2019)Ginart, Naumov, Mudigere, Yang, and Zou]{ginart2019mixed}
Antonio Ginart, Maxim Naumov, Dheevatsa Mudigere, Jiyan Yang, and James Zou.
\newblock Mixed dimension embeddings with application to memory-efficient recommendation systems, 2019.

\bibitem[Glorot and Bengio(2010)]{Glorot2010UnderstandingTD}
Xavier Glorot and Yoshua Bengio.
\newblock Understanding the difficulty of training deep feedforward neural networks.
\newblock In \emph{AISTATS}, 2010.

\bibitem[Guo et~al.(2017)Guo, Tang, Ye, Li, and He]{Guo2017DeepFMAF}
Huifeng Guo, Ruiming Tang, Yunming Ye, Zhenguo Li, and Xiuqiang He.
\newblock Deepfm: A factorization-machine based neural network for ctr prediction.
\newblock In \emph{IJCAI}, 2017.

\bibitem[Guo et~al.(2020)Guo, Alvarez, and Salzmann]{guo2020expandnets}
Shuxuan Guo, Jose~M Alvarez, and Mathieu Salzmann.
\newblock Expandnets: Linear over-parameterization to train compact convolutional networks.
\newblock \emph{Advances in Neural Information Processing Systems}, 33:\penalty0 1298--1310, 2020.

\bibitem[He and Chua(2017)]{he2017NFM}
Xiangnan He and Tat-Seng Chua.
\newblock Neural factorization machines for sparse predictive analytics.
\newblock In \emph{Proceedings of the 40th International ACM SIGIR Conference on Research and Development in Information Retrieval}, SIGIR '17, page 355–364, New York, NY, USA, 2017. Association for Computing Machinery.
\newblock ISBN 9781450350228.
\newblock \doi{10.1145/3077136.3080777}.
\newblock URL \url{https://doi.org/10.1145/3077136.3080777}.

\bibitem[Kaggle(2014)]{AvazuKaggle2014}
Kaggle.
\newblock Avazu click-through rate prediction, 2014.
\newblock \url{https://www.kaggle.com/c/avazu-ctr-prediction}.

\bibitem[Khrulkov et~al.(2019)Khrulkov, Hrinchuk, Mirvakhabova, and Oseledets]{khrulkov2019tensorized}
Valentin Khrulkov, Oleksii Hrinchuk, Leyla Mirvakhabova, and Ivan Oseledets.
\newblock Tensorized embedding layers for efficient model compression.
\newblock \emph{arXiv preprint arXiv:1901.10787}, 2019.

\bibitem[Krishnamoorthi(2018)]{Krishnamoorthi2018QuantizingDC}
Raghuraman Krishnamoorthi.
\newblock Quantizing deep convolutional networks for efficient inference: A whitepaper.
\newblock \emph{ArXiv}, abs/1806.08342, 2018.

\bibitem[Labs(2014)]{CriteoKaggle2014}
Criteo Labs.
\newblock Kaggle display advertising challenge dataset, 2014.
\newblock \url{http://labs.criteo.com/2014/02/kaggle-display-advertising-challenge-dataset/}.

\bibitem[Lian et~al.(2018)Lian, Zhou, Zhang, Chen, Xie, and Sun]{Lian2018xDeepFMCE}
Jianxun Lian, Xiaohuan Zhou, Fuzheng Zhang, Zhongxia Chen, Xing Xie, and Guangzhong Sun.
\newblock xdeepfm: Combining explicit and implicit feature interactions for recommender systems.
\newblock \emph{Proceedings of the 24th ACM SIGKDD International Conference on Knowledge Discovery \& Data Mining}, 2018.

\bibitem[Ling et~al.(2016)Ling, Song, and Roth]{ling-etal-2016-word}
Shaoshi Ling, Yangqiu Song, and Dan Roth.
\newblock Word embeddings with limited memory.
\newblock In \emph{Proceedings of the 54th Annual Meeting of the Association for Computational Linguistics (Volume 2: Short Papers)}, pages 387--392, Berlin, Germany, August 2016. Association for Computational Linguistics.
\newblock \doi{10.18653/v1/P16-2063}.
\newblock URL \url{https://www.aclweb.org/anthology/P16-2063}.

\bibitem[Naumov et~al.(2018)Naumov, Diril, Park, Ray, Jablonski, and Tulloch]{naumov2018periodic}
Maxim Naumov, Utku Diril, Jongsoo Park, Benjamin Ray, Jedrzej Jablonski, and Andrew Tulloch.
\newblock On periodic functions as regularizers for quantization of neural networks.
\newblock \emph{arXiv preprint arXiv:1811.09862}, 2018.

\bibitem[Naumov et~al.(2019)Naumov, Mudigere, Shi, Huang, Sundaraman, Park, Wang, Gupta, Wu, Azzolini, Dzhulgakov, Mallevich, Cherniavskii, Lu, Krishnamoorthi, Yu, Kondratenko, Pereira, Chen, Chen, Rao, Jia, Xiong, and Smelyanskiy]{Naumov2019DeepLR}
Maxim Naumov, Dheevatsa Mudigere, Hao-Jun~Michael Shi, Jianyu Huang, Narayanan Sundaraman, Jongsoo Park, Xiaodong Wang, Udit Gupta, Carole-Jean Wu, Alisson~G. Azzolini, Dmytro Dzhulgakov, Andrey Mallevich, Ilia Cherniavskii, Yinghai Lu, Raghuraman Krishnamoorthi, Ansha Yu, Volodymyr~Y. Kondratenko, Stephanie Pereira, Xianjie Chen, Wenlin Chen, Vijay Rao, Bill Jia, Liang Xiong, and Misha Smelyanskiy.
\newblock Deep learning recommendation model for personalization and recommendation systems.
\newblock \emph{ArXiv}, abs/1906.00091, 2019.

\bibitem[Ouyang et~al.(2019)Ouyang, Zhang, Ren, Li, Liu, and Du]{Ouyang2019ClickthroughRP}
Wentao Ouyang, Xiuwu Zhang, Shukui Ren, Linlin Li, Zhaojie Liu, and Y.~Du.
\newblock Click-through rate prediction with the user memory network.
\newblock \emph{ArXiv}, abs/1907.04667, 2019.

\bibitem[Ruder(2016)]{Ruder2016AnOO}
Sebastian Ruder.
\newblock An overview of gradient descent optimization algorithms.
\newblock \emph{ArXiv}, abs/1609.04747, 2016.

\bibitem[Shen(2019)]{shen2019deepctrtorch}
Weichen Shen.
\newblock Deepctr-torch: Easy-to-use,modular and extendible package of deep-learning based ctr models.
\newblock \url{https://github.com/shenweichen/DeepCTR-Torch}, 2019.

\bibitem[Shi et~al.(2019)Shi, Mudigere, Naumov, and Yang]{shi2019compositional}
Hao-Jun~Michael Shi, Dheevatsa Mudigere, Maxim Naumov, and Jiyan Yang.
\newblock Compositional embeddings using complementary partitions for memory-efficient recommendation systems.
\newblock \emph{arXiv preprint arXiv:1909.02107}, 2019.

\bibitem[Song et~al.(2019)Song, Shi, Xiao, Duan, Xu, Zhang, and Tang]{Song2019AutoIntAF}
Weiping Song, Chence Shi, Zhiping Xiao, Zhijian Duan, Yewen Xu, Ming Zhang, and Jian Tang.
\newblock Autoint: Automatic feature interaction learning via self-attentive neural networks.
\newblock In \emph{CIKM '19}, 2019.

\bibitem[Sun et~al.(2016)Sun, Guo, Lan, Xu, and Cheng]{sun2016sparse}
Fei Sun, Jiafeng Guo, Yanyan Lan, Jun Xu, and Xueqi Cheng.
\newblock Sparse word embeddings using l1 regularized online learning.
\newblock In \emph{Proceedings of the Twenty-Fifth International Joint Conference on Artificial Intelligence}, pages 2915--2921. AAAI Press, 2016.

\bibitem[Sutskever et~al.(2013)Sutskever, Martens, Dahl, and Hinton]{Sutskever2013OnTI}
Ilya Sutskever, James Martens, George~E. Dahl, and Geoffrey~E. Hinton.
\newblock On the importance of initialization and momentum in deep learning.
\newblock In \emph{International Conference on Machine Learning}, 2013.

\bibitem[Tissier et~al.(2019)Tissier, Gravier, and Habrard]{nearlossless_tissier}
Julien Tissier, Christophe Gravier, and Amaury Habrard.
\newblock Near-lossless binarization of word embeddings.
\newblock \emph{Proceedings of the AAAI Conference on Artificial Intelligence}, 33:\penalty0 7104–7111, Jul 2019.
\newblock ISSN 2159-5399.
\newblock \doi{10.1609/aaai.v33i01.33017104}.
\newblock URL \url{http://dx.doi.org/10.1609/aaai.v33i01.33017104}.

\bibitem[Wang et~al.(2017)Wang, Fu, Fu, and Wang]{Wang2017DeepC}
Ruoxi Wang, Bin Fu, Gang Fu, and Mingliang Wang.
\newblock Deep \& cross network for ad click predictions.
\newblock In \emph{ADKDD'17}, 2017.

\bibitem[Yang et~al.(2020)Yang, Huang, Park, Tang, and Tulloch]{yang2020mixedprecision}
Jie~Amy Yang, Jianyu Huang, Jongsoo Park, Ping Tak~Peter Tang, and Andrew Tulloch.
\newblock Mixed-precision embedding using a cache, 2020.

\bibitem[Yang and Zhang(2022)]{Yang2022ExpansionSqueezeBL}
Linzhuo Yang and Lan Zhang.
\newblock Expansion-squeeze- block: Linear over-parameterization with shortcut connections to train compact convolutional networks.
\newblock \emph{2022 8th International Conference on Big Data Computing and Communications (BigCom)}, pages 19--28, 2022.

\bibitem[Yin et~al.(2021)Yin, Acun, Liu, and Wu]{Yin2021TTRecTT}
Chunxing Yin, Bilge Acun, Xing Liu, and Carole-Jean Wu.
\newblock Tt-rec: Tensor train compression for deep learning recommendation models.
\newblock \emph{ArXiv}, abs/2101.11714, 2021.

\end{thebibliography}

\appendix





\end{document}